\documentclass[10pt,twocolumn,letterpaper]{article}

\usepackage{cvpr}
\usepackage{times}
\usepackage{epsfig}
\usepackage{graphicx}
\usepackage{amsmath}
\usepackage{amssymb}

\usepackage[breaklinks=true,bookmarks=false]{hyperref}

\cvprfinalcopy 

\def\cvprPaperID{2869} 

\begin{document}

\title{Learning Selfie-Friendly Abstraction from Artistic Style Images\thanks{Code available at: \url{https://github.com/DandilionLau/Selfie-Friendly-Abstraction}}}

\author{Yicun Liu \hspace{0.1in} Jimmy Ren \hspace{0.1in} Jianbo Liu \hspace{0.1in} Jiawei Zhang \hspace{0.1in} Xiaohao Chen \hspace{0.1in} \\
SenseTime Research\thanks{Demo available at: \url{https://youtu.be/0AsY26MHih4}}\\
\tt\small \{liuyicun,rensijie,liujianbo,zhangjiawei,chenxiaohao\}@sensetime.com
}

\maketitle

\begin{abstract}
Artistic style transfer can be thought as a process to generate different versions of abstraction of the original image. However, most of the artistic style transfer operators are not optimized for human faces thus mainly suffers from two undesirable features when applying them to selfies. First, the edges of human faces may unpleasantly deviate from the ones in the original image. Second, the skin color is far from faithful to the original one which is usually problematic in producing quality selfies. In this paper, we take a different approach and formulate this abstraction process as a gradient domain learning problem. We aim to learn a type of abstraction which not only achieves the specified artistic style but also circumvents the two aforementioned drawbacks thus highly applicable to selfie photography. We also show that our method can be directly generalized to videos with high inter-frame consistency. Our method is also robust to non-selfie images, and the generalization to various kinds of real-life scenes is discussed. We will make our code publicly available.
\end{abstract}

\begin{figure*}[t]
\begin{center}
\centerline{\includegraphics[width=1\linewidth]{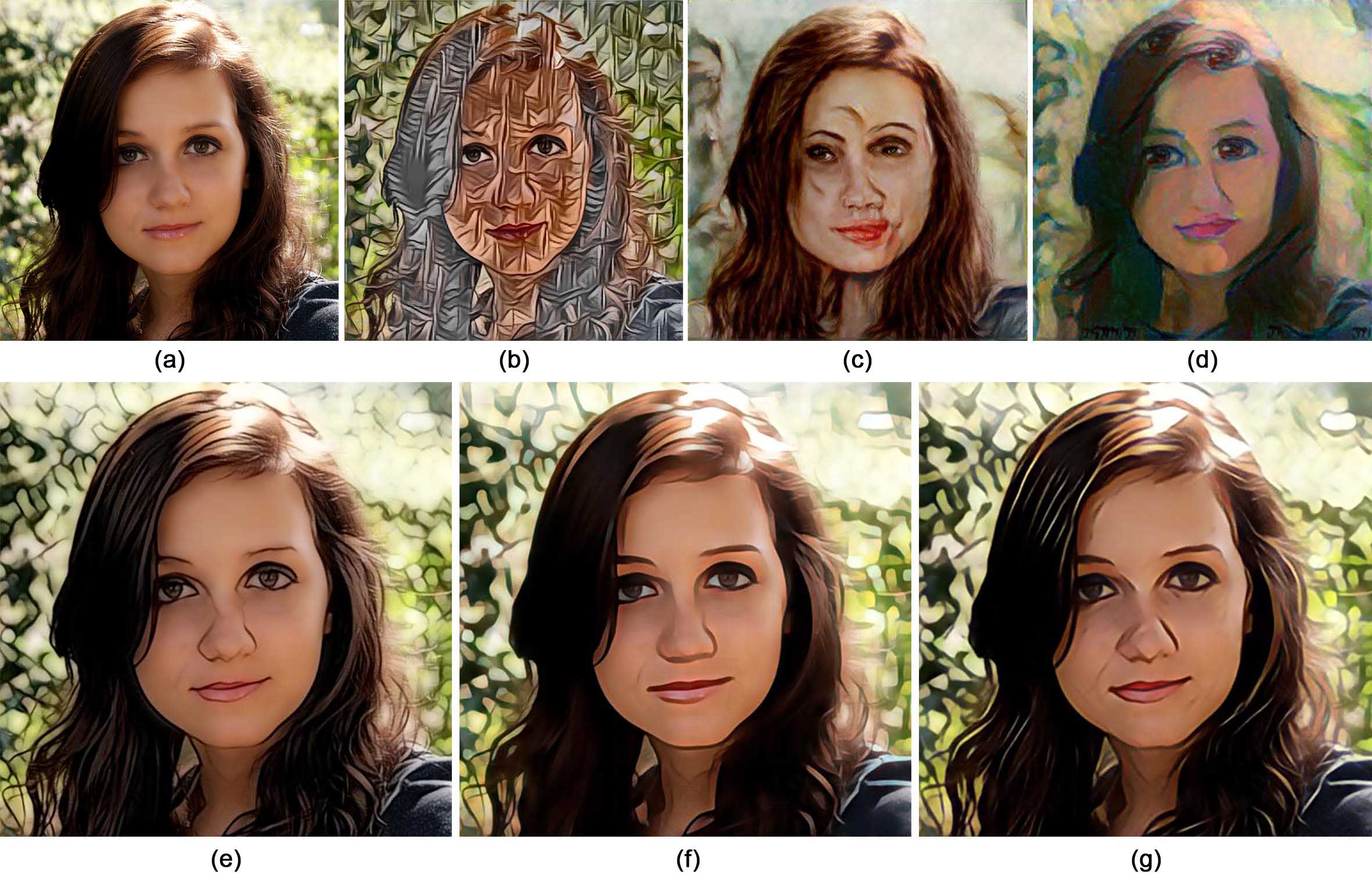}}
\caption{Even if with color preservation constraints, many style generation algorithms still perform poorly on facial images. In our experiment, (a) is the original image, (b) is Fast Neural Style \cite{Johnson2016} with luminance-only transfer \cite{Gatys2016b} to preserve the original color, (c) is Deep Analogy \cite{Liao2017}, and (d) is CNNMRF \cite{Li_2016_CVPR}. The style references for (c) and (d) is carefully selected similar female portrait paintings. Although these methods work fine in highly-abstracted style, they suffer from color shift and shape deformation when applied in fine-grained human facial cases. As comparison, image (e)(f)(g) are our results with different gradient-domain style.}
\label{fig:1}
\end{center}
\vskip -0.3in
\end{figure*}

\section{Introduction}
In the art world of painting creation, realistic traits of real-world scenes are often represented by a variety of artistic abstractions. The traditional filtering method of image abstraction usually smoothed the image while preserving various levels of structure. Recently, with the rise of deep learning, a seminal method for style generation has been proposed by \cite{Gatys2015}.

So far, despite various methods of image abstraction and style generation have been explored, precisely depicting human faces in artistic style remains challenging due to the strait restrictions on structural realism and color consistency. On the one hand, the human visual system is incredibly sensitive to irregularities in faces \cite{Jing2017,Selim2016}, even minor deformation in edges will affect the accuracy of human facial identification, leading to the unrealistic feeling of the poorly stylized version of human faces. On the other hand, because skin tone mainly serves as an essential visual feature for human faces, maintaining skin color in the stylistic version of selfies is crucial \cite{Shih2014}. 

In this paper, we aim to learn the `selfie-friendly' abstraction to both precisely and vividly depict human faces in artistic styles. To this end, we proposed a selfie-optimized CNN with a gradient domain training procedure. Unlike previous schemes, our method aims at learning the style abstraction directly on the gradient domain of images and can well tackle the two aforementioned drawbacks. Our framework is capable of learning different artistic abstractions while preserving the structural and color information in the original selfie. Another benefit of this innovation is that the framework can be directly used to render artistic style videos, with no flicking effect and convincingly high inter-frame consistency. 

\begin{figure*}[t]
\begin{center}
\vskip -0.2in
\centerline{\includegraphics[width=1\linewidth]{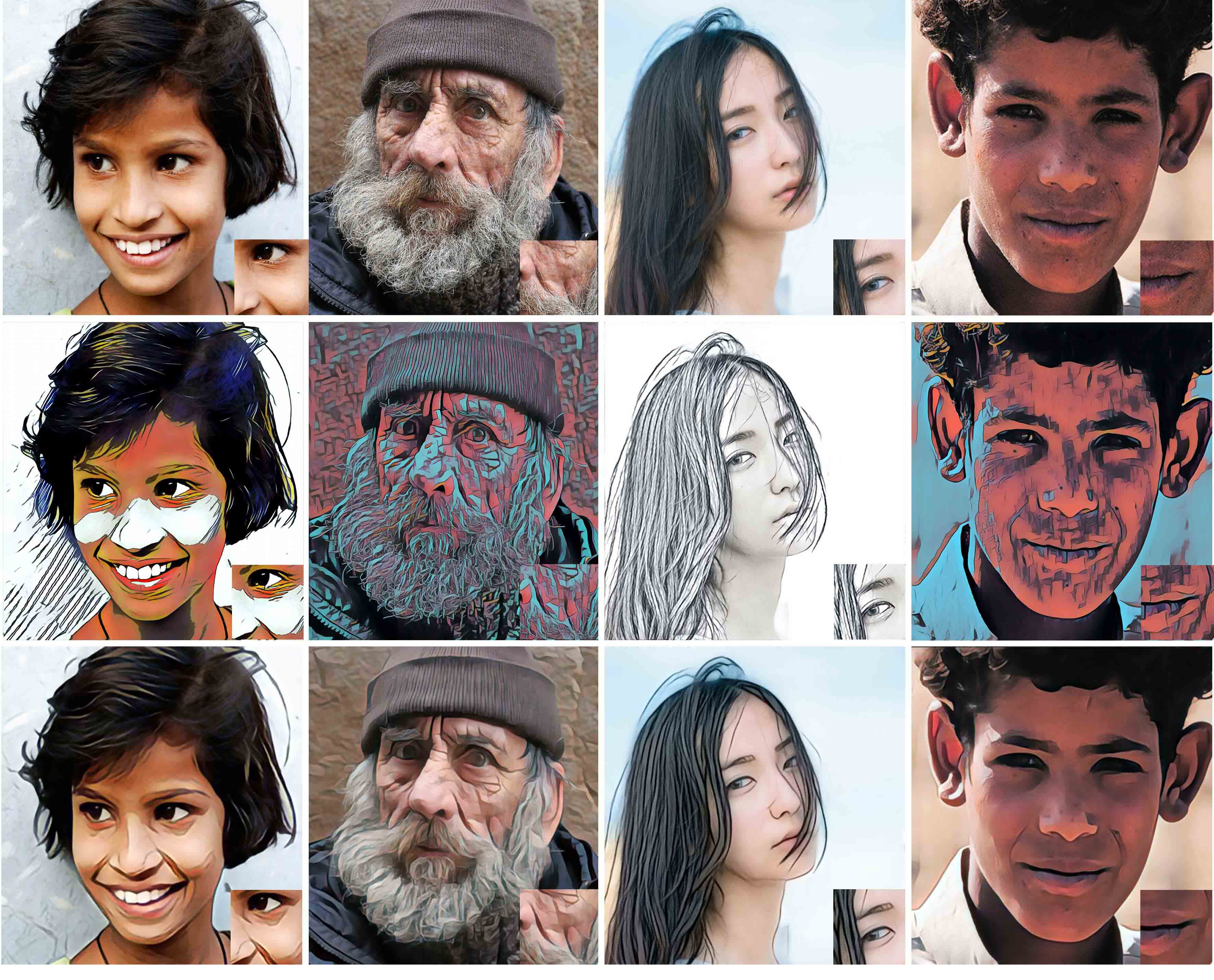}}
\caption{Selfie-friendly results generated by our algorithm: The first row are the original selfie photos. The second row are non selfie friendly references generated from existing method, which often come with severe distortion and color blemish. The third row are results generated from our method, learned from style sources in the second column. Please zoom in on monitor for better comparison.}
\label{fig:2}
\end{center}
\vskip -0.3in
\end{figure*}

To ensure the extent of style abstraction in sophisticated cases, we are the first to show that using the perceptual loss in the gradient domain can capture the stylistic traits of artistic images. Furthermore, the application of our framework is not limited to the selfie images. Thanks to the nature of gradient processing, it can also be generalized to diverse real-life scenarios which require color realism and structural consistency.

The main contribution of our work can be summarized in the following perspectives:
\vspace{-0.3\baselineskip}
\begin{itemize} 
\item[$\bullet$] First, we investigate two critical drawbacks of previous style generation methods on human faces and propose our selfies-friendly style abstraction framework that fully circumvents these drawbacks and achieves more attractive results for facial image stylization.
\vspace{-0.4\baselineskip}
\item[$\bullet$] Second, we explore the potential of gradient domain learning in the task of style abstraction. Our novelty includes applying firstly perceptual loss directly on gradient domain and color recovery from gradient images to comprehensively retain the original skin color.
\vspace{-0.4\baselineskip}
\item[$\bullet$] Third, our method tackles the drawback of flicking effect in style videos. In the video stylization task, it manifests high inter-frame consistency and is capable of rendering flicking-free artistic style videos, which is highly applicable to daily video streaming. 
\end{itemize}

\begin{figure*}[t]
\begin{center}
\vskip -0.3in
\centerline{\includegraphics[width=1\linewidth]{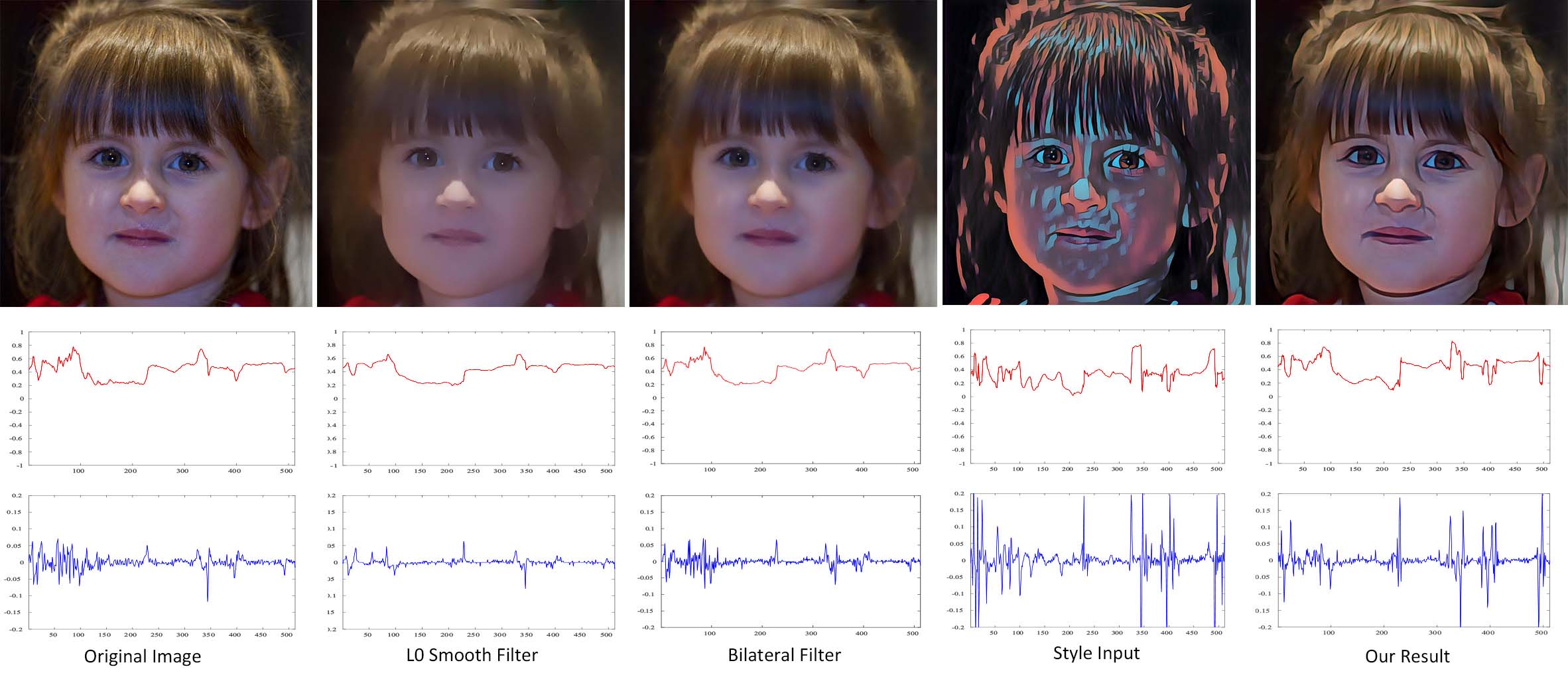}}
\caption{Color and gradient domain analysis: figures in the second row are the sectional views of the red channel, and figures in the third row are the sectional views of the horizontal gradient map. These two edge-aware filters smooth out most details except salient edges, with the variation being lower in both color and gradient domain. It is worth notice that the gradient view of our style input extremely diverts from the gradient views of the original image and edge-aware filter output, which is mainly resulted from distortions and color shift. Although in color domain, the structural information is usually mixed with chromatic information, the distinction is much more clear in gradient domain. In the training process, our method learns the edge-preserved style abstraction from candidate gradient patches. To this end, it is possible to diminish those defective scales and cracks and circumvent the visual drawbacks appearing in the style input. }
\label{fig:3}
\end{center}
\vskip -0.1in
\end{figure*}

\section{Related Work}
Artistic style abstraction and style generation have always been an open-ended challenge. Painting creation itself can be thought of a combination of the two tasks. Previously, image abstraction was investigated by many image processing papers. Many of the traditional filter-based methods handle image abstraction in an edge-aware manner, aiming to manipulate the rest of the image while preserving key structural information like edges. They have received a great deal of attention, like local Laplacian filter \cite{Paris2015}, L0 smooth filter \cite{Xu2011}, rolling guidance filter \cite{Qi2014}, and bilateral grid processing \cite{Chen2007}. Those methods demonstrate satisfying results in image abstraction, but unable to deal with complex abstraction tasks like artistic style abstraction, which requires semantic information of the image rather than local filter processing.

As the other task, style generation was usually decomposed into multi-levels: brushstroke level \cite{Lu2010}, texture level \cite{Efros2001} and patch level \cite{Meng2010}. Recently, CNN based style generation algorithm like \cite{Gatys2015,Johnson2016,Liao2017} aimed to learn the in-depth representation of style images, achieving impressive results. Meanwhile, many creative application tasks like learning filter banks for style generation \cite{chen2017stylebank} and style transfer on stereoscopic view \cite{chen2018stereoscopic} are extensively studied. Most style generation techniques suffer from unwanted defects in facial depiction, ranging from deformation in facial edges to severe color shift in skin tone. Even with color preservation like luminance-only transfer \cite{Gatys2016b} to correct the skin color, there still exists severe blemish. An example is showed in figure \ref{fig:1}.  

Few published papers addressed the problem of unsatisfying defects in skin color and facial edges \cite{Selim2016,Sucontphunt2014}. Only limited amount of authors have noticed similar irregularities when they dealt with facial images. Their attempts included reducing portrait distortion by exploiting human face geometry \cite{Chen2001}. and recovering skin smoothness from brushstroke transfer task \cite{Regan2009}. However, these approaches were used required intensive computational cost and were only applicable to limited styles like sketches style. There were also attempts to recover the original color in style transformation \cite{Gatys_2017_CVPR,Luan_2017_CVPR}, with some methods using color histogram matching and luminance-only transfer \cite{Gatys2016b}. Generally, their approaches suffered from extra tradeoff in the extent of stylization and were not specially designed for selfies. Many video-oriented stylization methods have been introduced \cite{Huang2017,Chen_2017_ICCV}, but only a few were concentrating on maintaining inter-frame consistency in style transfer task has also been studied in \cite{Ruder2017}.

\begin{figure*}[t]
\begin{center}
\vskip -0.2in
\centerline{\includegraphics[width=1\linewidth]{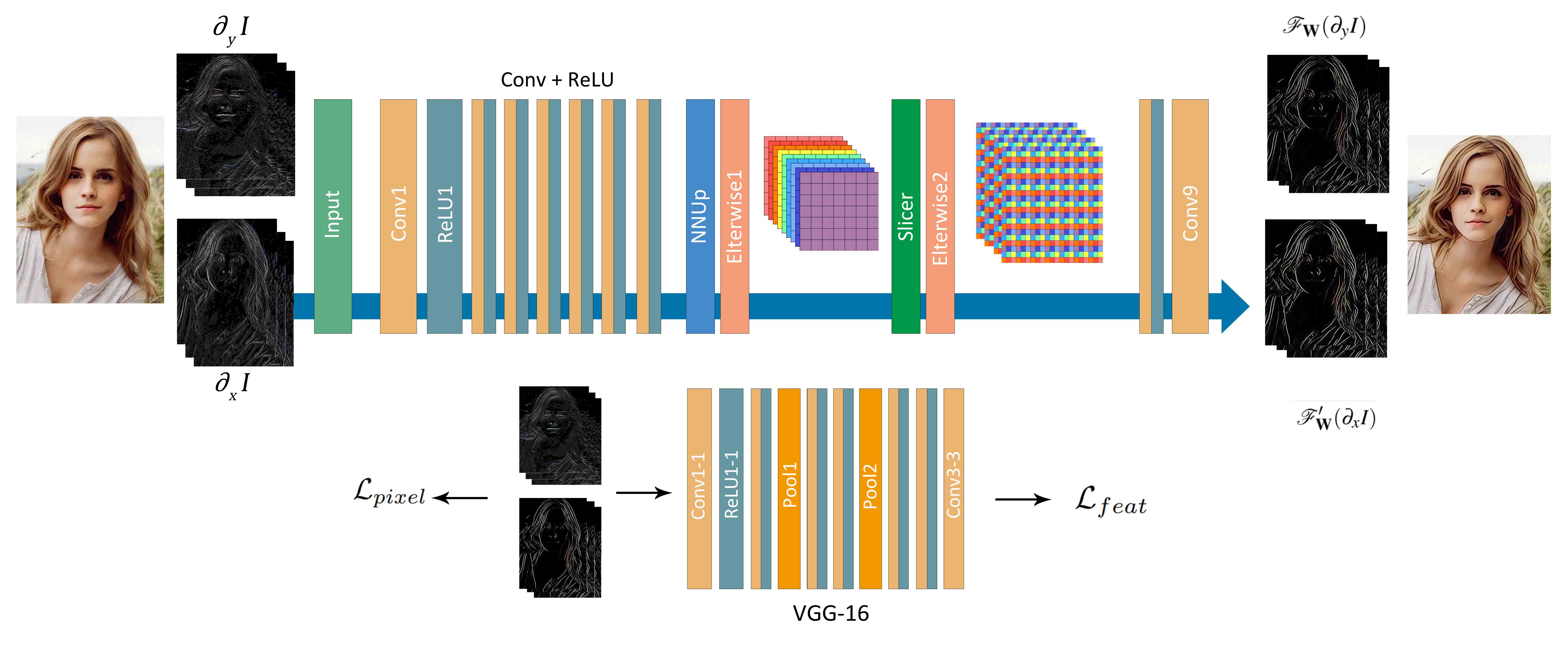}}
\caption{Architecture of our selfie-friendly stylization network. Our design operates on the gradient domain of input and reference images. Our network includes seven convolutional layers with ReLU to extract gradient features from the input images and the style references. Note that the height and width of the feature map in Conv-7 are downsampled by the scale factor of four. We later adopt the sub-pixel module to upsample the stylistic results to be the same as the original size. We calculate the pixel-wise loss $\mathcal{L}_{pixel}$ with the outputs of Conv-9 and style references. Then we used VGG-16 to calculate perceptual loss $\mathcal{L}_{feat}$ on the gradient domain.}
\label{fig:4}
\end{center}
\vskip -0.3in
\end{figure*}

The most proximate work to our goal is \cite{Selim2016,Shih2014}. The algorithm proposed by Slim aimed to reduce the distortion and increase visual fidelity by adding an example-driven spatial constraint for each facial image. One limitation of the example-based method is that it requires similar portrait images as the reference, which is hard to find. Their model also lacked global consideration for skin color consistency. Unlike \cite{Selim2016}, our method can be generalized to other relevant scenes and exhibits overall concern for multi-aspect naturalness preservation in the universal light-weighted framework.

Before our experiment, the topic of gradient domain processing has been merely explored since the rise of deep neural network \cite{Hinton2014}. The potential of using gradient training in deep convolutional neural network has still not been throughout looked into. Early work mainly focused on gradient tone mapping and gradient dynamic range compression \cite{Socolinsky2012}. Xu is the first to employed gradient domain training to accelerate deep edge-aware filters, their results showed promising capabilities in preserving structural realism \cite{xub15}. Gradient processing was also utilized in \cite{Mechrez2017} for photorealistic style transfer. Nevertheless, there is still no further examination for using gradient domain training in specific structure-color-concentrating tasks, such as style abstraction and generation for selfies.

\section{Our Approach}
Our method takes two images: an input image $\textit{I}$ which is usually an ordinary selfie image and a corresponding stylized reference of the original selfie $\mathcal{L}\textit{(I)}$. $\mathcal{L}$ could be an existing style transfer algorithm, which usually generates reference stylized selfie images with apparent drawbacks of inaccurate edges and color shift. In this algorithm, we seek the optimized facial transformation $\mathcal{F}$ which transfers selfie image $\textit{I}$ to a visually more satisfying stylized version $\mathcal{F}_{\textbf{W}}\textit{(I)}$. Here $\mathcal{F}$ denotes the network architecture, and $\textbf{W}$ represents the network parameters. We name this optimized transformation as selfie-friendly transformation, which should not only generate the artistic abstraction of the reference but also avoid the aforementioned drawbacks. 

Our approach achieves selfie-friendly artistic style abstraction by introducing three core ideas to the traditional CNN based method:
\vskip -0.05in
\begin{itemize}
\item[$\bullet$] We propose a neural style architecture that is fully based on the gradient of images. The edge-aware nature of gradient learning provides constraints on edges to eliminating various distortions. 
\vskip -0.05in
\item[$\bullet$] We introduce color confidence in the reconstruction part to maintain the visual fidelity of the skin color in result images. The reconstruction step exploits both the structure and color information of the original selfies, which ensures the naturalness of the result.
\vskip -0.05in
\item[$\bullet$] We initiate the exploration of using perceptual loss directly on gradient domain to enhance the extent of stylization when learning abstraction of diverse style.
\end{itemize}

\subsection{Gradient Constraints and Objectives}

One simple strategy to learn the artistic abstraction is that to train the neural network by directly minimizing the summed up pixel-wise loss in RGB channels.
\begin{equation}\label{eq:1}
\left \|  \mathcal{F}_{\textbf{W}}\textit{(I)} - \mathcal{L} \right \|^2
\end{equation}

However, this attempt to maximize objective function directly on the color domain will inevitably lead to the problem of insensitivity to the edge structures. One example of the problems is shown in figure \ref{fig:1}, where the shape of the human face in the stylized results significantly diverts from the original selfie images. The skin color is also poorly represented and without smoothness and naturalness. In our further analysis of training in the color domain and gradient domain, we find that the gradient of stylistic reference images meaningfully diverts from the gradient of the original images, as shown in figure \ref{fig:3}. 

In order to better evaluate the significance of training in gradient domain, we adopt the biliteral edge-aware filter proposed in \cite{xub15}. The edge-aware filter smoothes out most of the detailed structure but preserves important edges. Another observation shows that in most of the previous defective samples, the deformation and color shift problem occurs nondeterministically. For example, selfies of the same person taken from slightly different angles can lead to very different facial edges in stylistic patches. In the training process, our method mostly learns those edge-preserved abstractions of the candidate patches in the gradient domain. It is possible to diminish the effect of the defective patches in gradient domain and circumvent the visual drawbacks appearing in previous examples.

With the above understanding, we define the objective function on $\nabla\textit{I}$ rather than $\textit{I}$. Considering that most edge-aware operators can produce the same effects even if we rotate the input image by 90 degrees, we use both the vertical gradient $\nabla\textit{I}_{\textbf{V}}$ and horizontal gradient $\nabla\textit{I}_{\textbf{H}}$ in our training process. Here we denote $\partial\textit{I}$ as the channel-wise combination of vertical gradient $\nabla\textit{I}_{\textbf{V}}$ and horizontal gradient $\nabla\textit{I}_{\textbf{H}}$.

Now given $D$ training image pairs ($\textit{I}_{0}$, $\mathcal{L}$\textit{($I_{0}$)}), ($\textit{I}_{1}$, $\mathcal{L}$\textit{($I_{1}$)}), $\cdots$, ($\textit{I}_{D-1}$, $\mathcal{L}$\textit{($I_{D-1}$)}) of the original selfie images and corresponding unsatisfying stylized reference images, we aim to minimize
\begin{equation}\label{eq:2}
\dfrac{1}{D}\sum_{i} \left \{\dfrac{1}{2}\left \|  \mathcal{F}_{\textbf{W}}(\partial\textit{I}_{i}) - \partial\mathcal{L}(\textit{I}_{i})\right \|^2)\right \}
\end{equation}
where $\left \{\partial\textit{I}_{i}, \partial\mathcal{L}(\textit{I}_{i}) \right \}$ denotes the traning example pair in gradient domain. By minimizing the objective function in gradient domain,  structural content includes facial edges can be specially emphasized and carefully preserved in the traning process of our framwork. 

\begin{figure}[t]
\begin{center}
\vskip -0.1in
\centerline{\includegraphics[width=1\linewidth]{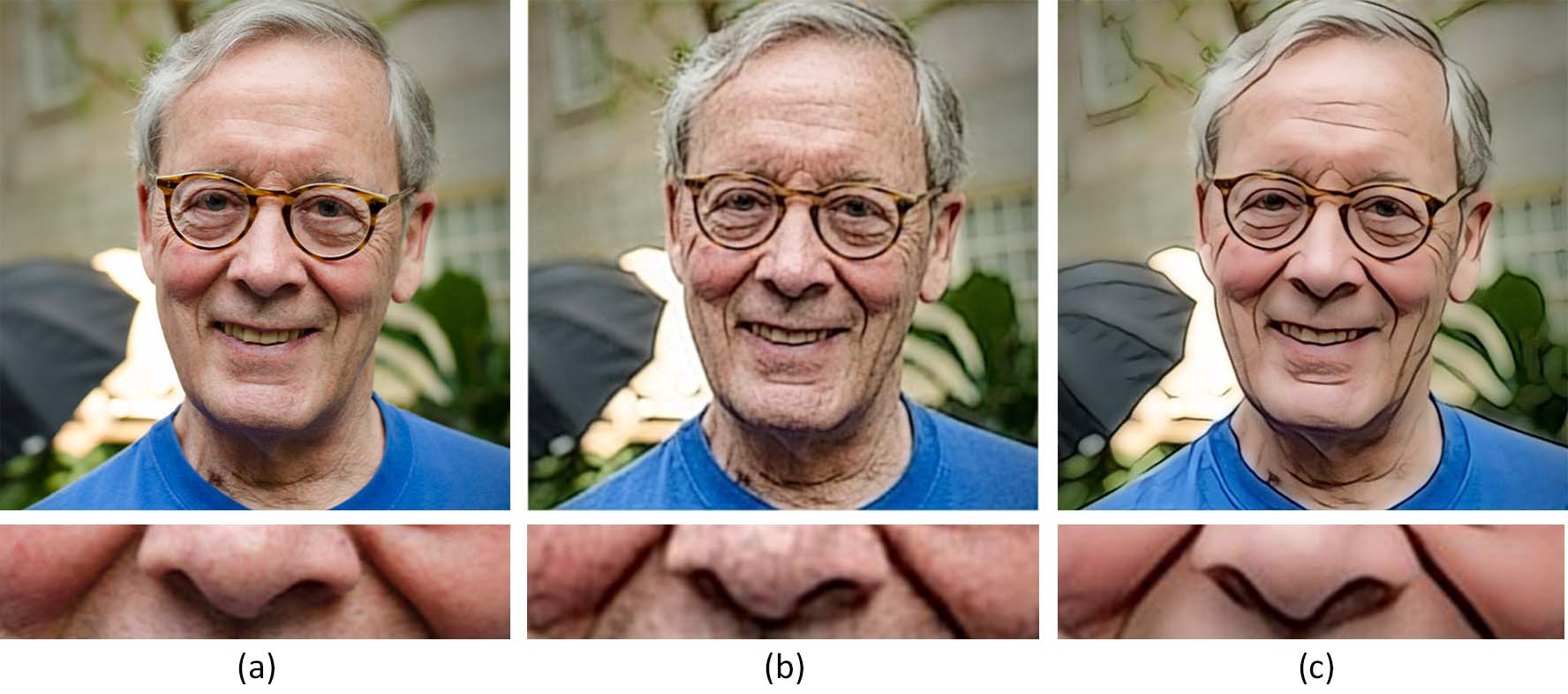}}
\caption{Perceptual loss on gradient domain: (a) The original selfie photo (b) Training result with only pixel loss (c) Training result with both pixel loss and perceptual loss. Without perceptual loss applied on gradient domain, stylization becomes less abstract and more defective as shown in (b), in which the skin smoothness and stylization quality is not as satisfying as (c).  }
\label{fig:5}
\end{center}
\vskip -0.35in
\end{figure}

\begin{figure*}[t]
\begin{center}
\vskip -0.2in
\centerline{\includegraphics[width=1\linewidth]{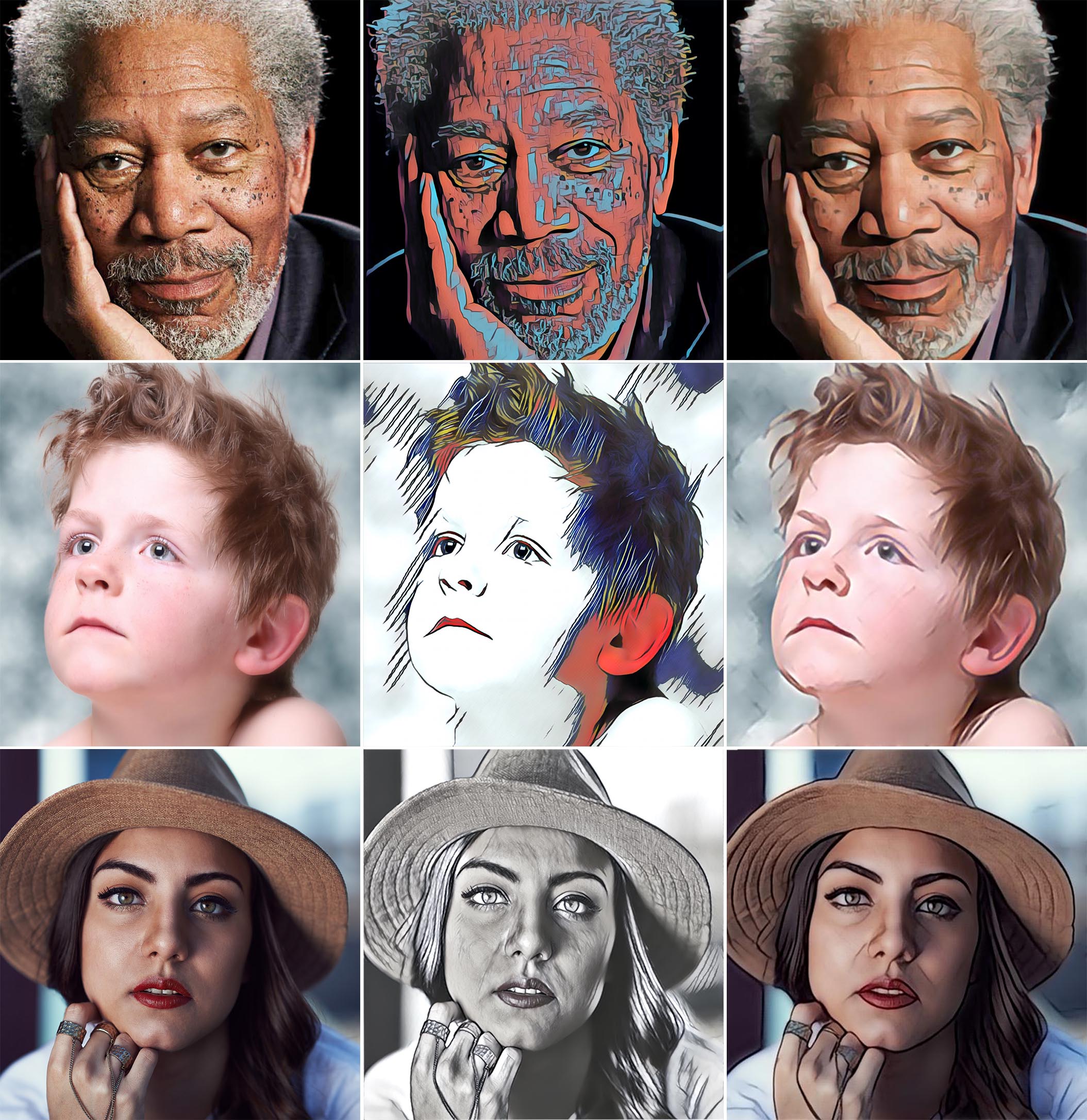}}
\caption{Experiments on people of different age, gender and race: The first column are the original selfie images, and the second column are the stylistic references. The third column are our results. Please zoom in on monitor for better comparison.}
\label{fig:6}
\end{center}
\vskip -0.35in
\end{figure*}

\begin{figure*}[t]
\vskip -0.2in
\begin{center}
\centerline{\includegraphics[width=1\linewidth]{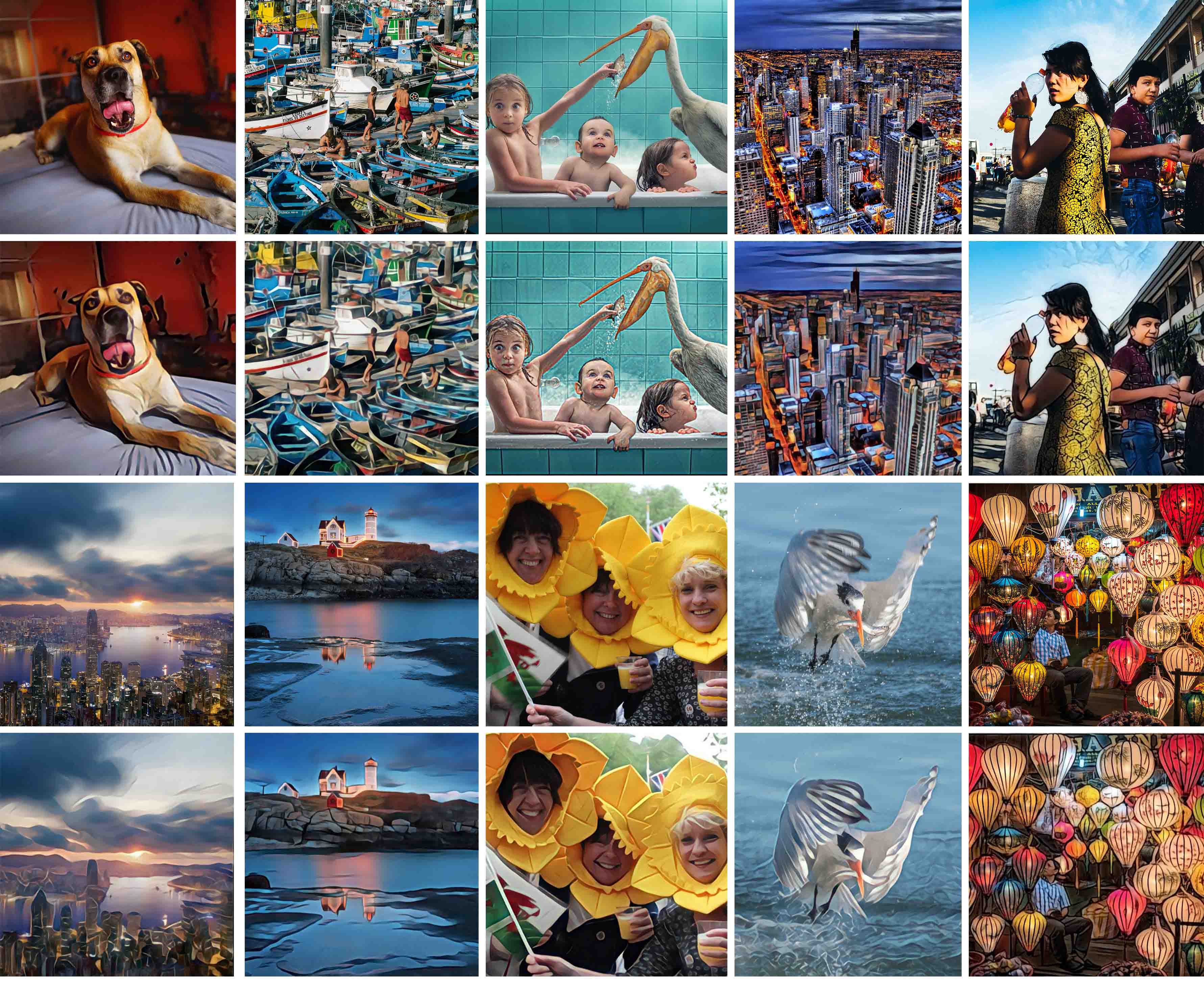}}
\caption{Experiments on non-facial scenarios. Note that the training set is totally based on facial images and our method has impressive generalization capability for all cases in which structural realism and color consistency are considered. }
\label{fig:7}
\end{center}
\vskip -0.35in
\end{figure*}

\subsection{Perceptual Loss on Gradient Domain}
We make an attempt to apply perceptual loss directly on gradient domain, based on our finding that only using the pixel-wise loss on gradient domain can be restrictive. Perceptual loss enhances the stylization process by extracting high-level semantic representations of edges. Those edges can be important information like brushstrokes in paintings. It is proved that gradient-level semantic information is quite meaningful for our task. An illustration of result with and without the perceptual loss is shown in figure \ref{fig:5}. 

Following the concept at \cite{Johnson2016}, we define the perceptual loss based on VGG-16 image classification network pretrained on ImageNet. The preceptual loss is defined as the $\textit{l}_{2}$-norm between feature representations of the reference image $\textit{I}$ and the output stylized image $\mathcal{F}_{\textbf{W}}\textit{(I)}$.
\begin{equation}\label{eq:3}
\mathcal{L}_{feat} = \dfrac{1}{\mathcal{C}_{j}\mathcal{H}_{j}\mathcal{W}_{j}}\left \|\psi_{j}(\textit{I}) - \psi_{j}(\mathcal{F}_{\textbf{W}}\textit{(I)})\right \|
\end{equation}
where $\psi_{j}()$ denotes the feature map from the \textit{j}-th VGG-16 convolutional layer and $\mathcal{C}_{j}$, $\mathcal{H}_{j}$, $\mathcal{W}_{j}$ are the number, height and width of the feature maps, respectively. Here we only use the conv3-3 layer for the final output of style representation. We calculate perceptual loss from the euclidean distance between the two outputs of conv3-3 layer.

\subsection{Total Loss for Training}
We formulate the artistic abstraction learning objective function, combing all two loss components together.
\begin{equation}\label{eq:4}
\mathcal{L}_{total} = \alpha\mathcal{L}_{pixel} + \beta\mathcal{L}_{feat}
\end{equation}
where $\alpha$ and $\beta$ are the corresponding loss weights for pixel loss and style loss. 

\subsection{Network Architecture}
The overall network architecture is illustrated in figure \ref{fig:4}. The network is mainly constructed of two parts: the first part takes the gradients of selfie images as input, using continuous convolutional layers with ReLU as feature extraction, then calculate the pixel-wise loss after a sub-pixel upsampling operation. In the second part, the result of the first part will be passed to a VGG-16 network, together with gradients of the reference stylized images. Noted that the weights of the VGG-16 network are fixed in the whole training procedure. After calculating perceptual loss from the output of conv3-3 layer, total loss will be summed up based on the optimized loss weights.

In this network, our architecture also embeds several micro-designs to mitigate checkerboard artifacts. We implement the sub-pixel convolution layer proposed in \cite{Shi_2016_CVPR} instead of deconvolution layer to avoid those artifacts in uneven deconvolution. The sub-pixel upsampling module includes an NNU layer which duplicates the input $N\times{N}$ times by nearest neighbor strategy, with a masking operation to map the sub-pixels into the corresponding position in the final upsampled feature map.

\subsection{Training Details}
For the training data, we use a set of selfie images gathered from Flickr as the original images and their corresponding stylistic versions generated from Prisma as the style references. We randomly collect $64\times64$ patches from selfie images and the corresponding style patches. In our training process, the loss weight $\alpha$ and $\beta$ are set to 10000 and 10 respectively, to ensure a smooth decrease in both $\mathcal{L}_{pxiel}$ and $\mathcal{L}_{feat}$. 

The network was trained on \textit{Nvidia Titan} X GPU for 100K iterations using a batch size of 10. The learning rate was set to be $10^{-8}$ in first 50K iterations and $10^{-9}$ in the second 50K iterations. We use Adam algorithm to minimize the total loss shown in eq. (\ref{eq:4}). The training procedure took about two hours, and the experimental setup was identical in all of the experiments.

\subsection{Image Reconstruction}

We denote by $\textit{S}$ our final output gradient map. To maximize the color naturalness and structural realism of human faces in our output, the reconstruction step also exploits the structural information and color content in the input image to guide smoothing in gradient domain. We thus introduce two terms adding together as

\begin{equation}\label{5}
\left \|\textit{S} - \textit{I}\right \| + \lambda\left \{\left \|\partial_{x}\textit{S} - \mathcal{F}^{'}_{\textbf{W}}(\partial_{x}\textit{I}) \right \|^{2} + \left \|\partial_{y}\textit{S} - \mathcal{F}_{\textbf{W}}(\partial_{y}\textit{I}) \right \|^{2}\right \}
\end{equation}
where $\left \|\textit{S} - \textit{I}\right \|$ is the color confidence to use the input image to guide the smoothed image reconstruction. The second term is the common one to use the gradient result both in horizontal and vertical axis. $\lambda$ is another parameter balancing the two terms.
\vskip 0.01in
Note that $\lambda$  is the balancing factor of color information and structural information in the reconstruction step. This value varies for diverting scenarios of input. Here we perform a greedy search in the result images, which is applied to the testing set of selfie images. In the color recovery process of our experiments, the value of $\lambda$ is set to be 10.

\begin{figure}[t]
\vskip -0.1in
\begin{center}
\centerline{\includegraphics[width=1\linewidth]{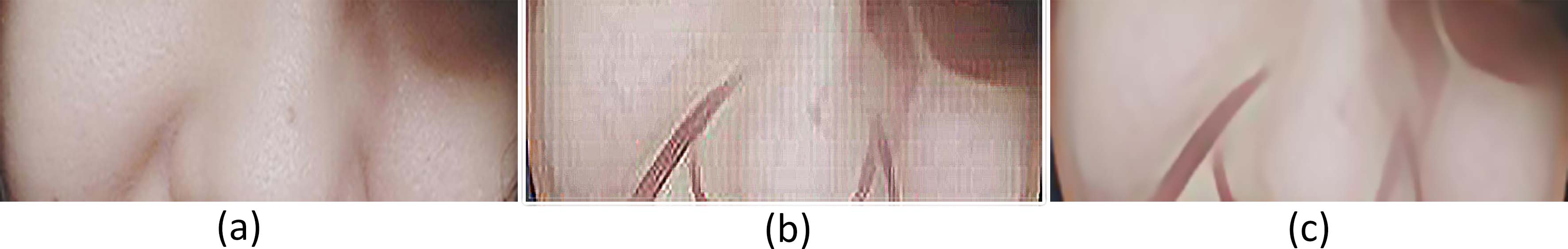}}
\caption{Our trail of removing checkerboard artifact: (a) The original photorealistic image (b) Training result with deconvolution layer (c) Training result with sub-pixel upsampling module. The checkerboard effects in (b) are eliminated in (c) since we adopt artifact-free architecture in our network setting.}
\label{fig:8}
\end{center}
\vskip -0.35in
\end{figure}

\section{Experiments and Applications}
We use our framework to learn the selfie-friendly artistic abstraction of a number of real-life stylization effects, including these popular effects generated by Prisma. Although their original algorithms vary in computational cost, our optimized selfie-friendly model can generate stylized images in a universal fast speed, while largely improving the visual attractiveness of the abstraction of facial features. The testing step takes images of size $512 \times 512$ as input, and each image takes on average 0.05 second to process using unoptimized MATLAB code.

\subsection{Visual Quality Assessment}
\begin{figure}[t]
\vskip -0.1in
\begin{center}
\centerline{\includegraphics[width=1\linewidth]{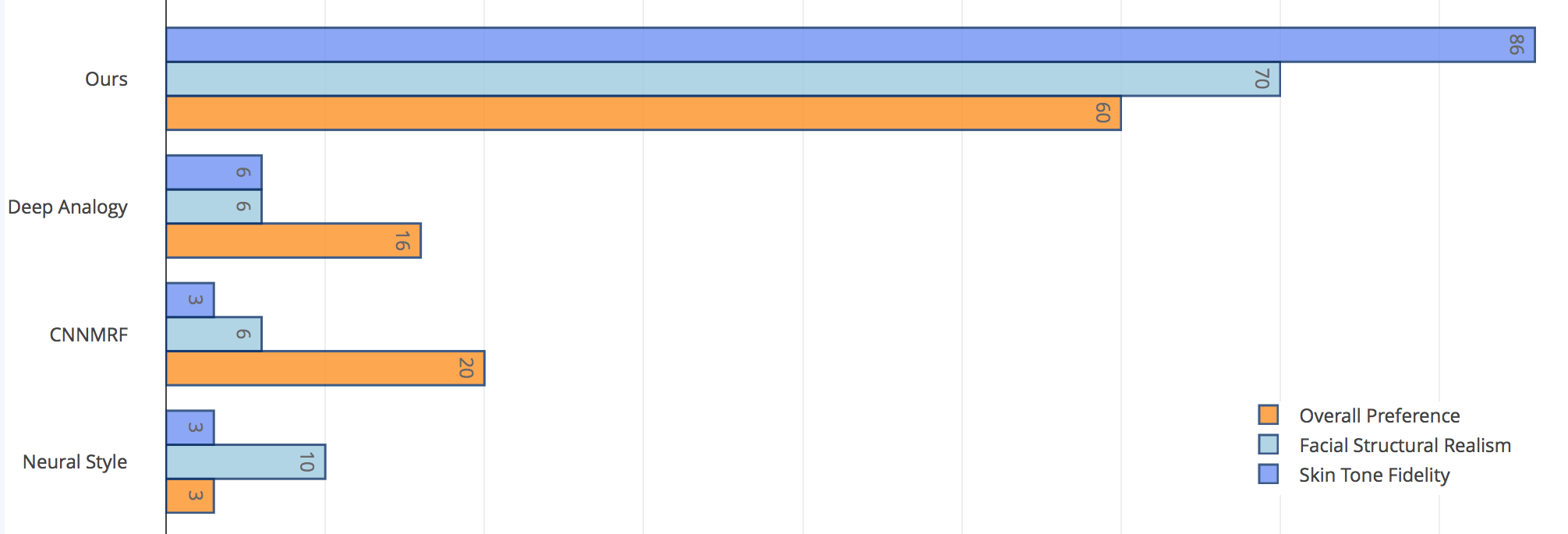}}
\caption{User survey confirming that our algorithm produces faithful results in both structural realism, skin tone fidelity and overall preference assessment. }
\label{fig:10}
\end{center}
\vskip -0.25in
\end{figure}

We conducted two user studies to validate our work. We assessed the results generated by our method, Deep Analogy \cite{Liao2017}, CNNMRF \cite{Li_2016_CVPR}, and Fast Neural Style \cite{Johnson2016}. To ensure a relatively fair assessment, we used color preservation technique proposed in \cite{Gatys2016b} for all other methods. The survey was conducted on a set of 20 selfie images and user needed to vote for the best method in terms of facial realism and overall preference. Our method won among those methods with more than 70\% of votes. The detail of the result is shown in figure \ref{fig:10}.

\begin{figure}[t]
\begin{center}
\vskip -0.1in
\centerline{\includegraphics[width=1\linewidth]{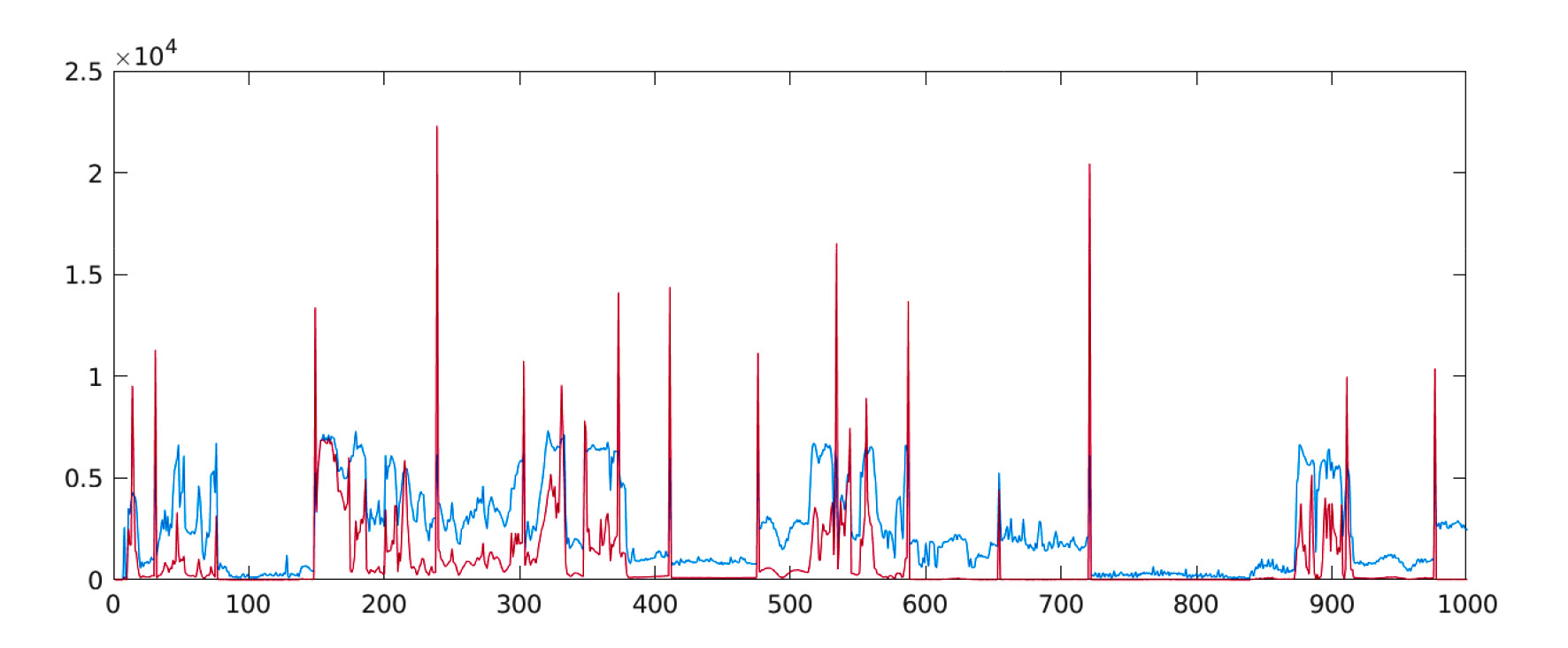}}
\caption{Inter-frame consistency test of our method (red) and the reference method (blue). The y axis is calculated by the MSE of continous frames. Our method demostrates higher consistency in multiple sample video footages. Note that the peaks in our method are related with scene change in the original footage.}
\label{fig:9}
\end{center}
\vskip -0.35in
\end{figure}

To ensure our method is robust for selfies in all circumstances (i.e. age, gender, race and etc.), we use selfies of people from diverse backgrounds. As shown in figure \ref{fig:6}, our experiment generates convincing results, with the output stylistic image preserving the substantial visual feature of different facial identities, showing that our framework can be adapted to various groups of people, regardless of facial feature differences arouse from their age, gender and race.

\subsection{Artifact Removal}
At the initial trial of our experiment, we simply use deconvolution layer to upsample the feature map in the first part of our network. The kernel size and stride length we used were also not delicately checked and some of the kernel size used is not dividable by stride in the same convolution layer. The initial result was dramatically affected by checkerboard artifacts, with bothering checkerboard artifacts, as shown in figure \ref{fig:8}. 

One reason is that overlapping occurs when kernel size is not divisible by stride, which often used as the upsampling factor in deconvolution operation \cite{Aitken2017,Odena2016}. After using the sub-pixel module in the first part of our network and carefully checking the kernel size all convolution layers, these annoying artifacts have been mostly eliminated in results generated from our framework.

\subsection{Style Generalization}
In this section, we use a large set of images downloaded from Flickr, including landscapes, buildings, and objects for generalization testing. Although the network is trained on selfie dataset, our model still demonstrates impressive generalization capability when dealing with various scenes. The result images embody salient style while preserving the original color and structure fidelity. Example of the results is shown in figure \ref{fig:7}.

\subsection{Extension to Videos}
To test the inter-frame consistency, we apply our model to several video clips of diverse contents, each of them contains over 5000 frames. Figure \ref{fig:9} compares the inter-frame consistency of our method and the most recent method in \cite{Ruder2017}. Different from existing methods which have severe effect of flicking when applied to videos, our method shows reliable inter-frame consistency. The stylistic videos generated are smooth and consistent, even for rapid inter-frame transitions.

\section{Conclusion}
In this paper, we proposed a novel artistic stylization framework that specially optimized for human facial images. The method we proposed exploits the structural information in the gradient domain to preserve structure realism, separates chromatic information from gradient information in the learning process. Our method uses color reconstruction to carefully preserve the skin tone of facial images. The gradient approach explicitly takes advantage of statistical properties in gradient domain to eliminate mismatched patches appearing in the style samples. From our experiments, we showed that our method can generate stylized selfie images qualified as more attractive in both visual and aesthetic assessments. 

{\small
\bibliographystyle{ieee}
\bibliography{egbib}
}

\end{document}